\documentclass[runningheads]{llncs}
\usepackage[T1]{fontenc}
\usepackage{mathtools}
\usepackage{amsfonts,amsmath,amssymb}
\usepackage{graphicx}
\usepackage{tikz}
\usetikzlibrary{positioning,arrows.meta}
\usetikzlibrary{calc}
\usepackage{booktabs}
\usepackage{multirow}
\begin{document}
\title{Real-time Localization of a Soccer Ball from a Single Camera}
%
%
\author{Dmitrii Vorobev\inst{1}\orcidID{0009-0005-2700-1243} \and
Artem Prosvetov\inst{1}\orcidID{0000-0001-7997-0920} \and
Karim Elhadji Daou\inst{2}\orcidID{0000-0003-4677-7571}}
\authorrunning{D. Vorobev et al.}
%
\institute{Yandex, Moscow, Russia \and Yandex, Almaty, Kazakhstan}
\maketitle              
\begin{abstract}
We propose a computationally efficient method for real-time three-dimensional football trajectory reconstruction from a single broadcast camera. In contrast to previous work, our approach introduces a multi-mode state model with $W$ discrete modes to significantly accelerate optimization while preserving centimeter-level accuracy -- even in cases of severe occlusion, motion blur, and complex backgrounds. The system operates on standard CPUs and achieves low latency suitable for live broadcast settings. Extensive evaluation on a proprietary dataset of 6K-resolution Russian Premier League matches demonstrates performance comparable to multi-camera systems, without the need for specialized or costly infrastructure. This work provides a practical method for accessible and accurate 3D ball tracking in professional football environments.

\end{abstract}

\section{Introduction}
\label{sec:intro_task}


Real-time three-dimensional localization of the football is becoming an essential tool in modern performance analysis, refereeing, and sports broadcasting.
At its core, the process involves detecting the ball in each video frame, inferring its XYZ coordinates relative to a global pitch reference from calibrated camera views (or depth priors), and filtering these estimates in real time for centimeter-level accuracy -- even in the presence of occlusion or motion blur.
Such precision enables advanced analytics, including metrics like expected goals, shot profiles, and pass sharpness, while supporting critical technologies like goal-line decisions and VAR that require objective evidence.
Accurate ball trajectories also allow broadcasters to add virtual offside lines, dynamic heatmaps, and other augmented-reality graphics, giving viewers a deeper insight into the team tactics.
Coaches, meanwhile, use this data to quantify ball tempo and spatial pressure, tying match execution to analytical feedback.
Robust tracking overcomes long-standing challenges from occlusion and blur in video, making automation viable across diverse stadium and camera setups.
Ultimately, precise ball tracking drives richer game analysis and content creation for more engaging fan experiences, making it a cornerstone of contemporary football technology.

However, challenges remain. In a 6K video frame, the ball covers less than \(0.15\%\) of the area and is often camouflaged by complex backgrounds. About half the time, it is completely hidden from the camera by players, goalposts, or referees, and its path rarely follows a simple parabola -- being affected by deflections, wet turf, and mis-hits. While multi-camera setups with global optimization offer highly accurate results, they require expensive infrastructure and specialized staff, costing hundreds of thousands of dollars per stadium.
In contrast, the main broadcast camera feed is universally available and incurs no extra cost, but extracting depth information from a single viewpoint is fundamentally challenging, making direct triangulation impossible.

Our objective is to develop an efficient real-time method to reconstruct the ball’s 3D trajectory from a single broadcast view with an accuracy comparable to that of multi-camera systems. Our solution runs entirely on standard CPUs and produces results in less than two seconds, meeting the low-latency requirements of live broadcasting.

\begin{figure}[ht]
    \centering
    \input{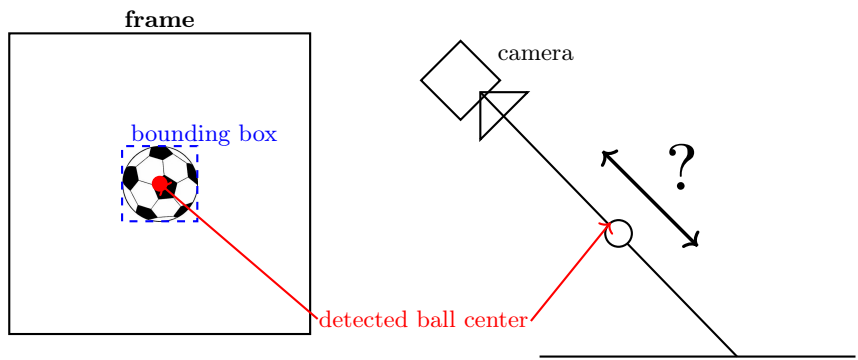}
    \caption{Uncertainty in ball position determined from only a single camera frame.}
    \label{ray1}
\end{figure}

Single-camera trajectory reconstruction differs fundamentally from the multi-camera case. With just one camera angle, it is practically unfeasible to determine the ball’s real world 3D position using only its image-plane coordinates, since the distance from camera to ball is unknown (see Fig.~\ref{ray1}). While some prior attempts \cite{van20223d} have tackled this problem, accuracy remains inadequate.

To address these challenges, we developed a multi-mode state model that distinguishes phases such as free flight, player possession, post-kick buffering, and out-of-play. The method applies limited discretization and a fixed-width beam search, ensuring linear time complexity for each filter step. We introduce a controllable latency parameter~\(L\), that allows dynamic trade-offs between accuracy and latency. We validate our solution on an in-house dataset of Russian Premier League matches (season 2024/25, 6K resolution), demonstrating real-time performance (50+ fps on a single Intel-Xeon-CPU-E5-2660 core) and an accuracy on par with offline optimization when \(L=50\) frames.

In summary, the key contributions of our paper are: a multi-mode state model, a linear-complexity optimization algorithm, and empirical validation on a real-world dataset. 

This work provides a practical framework for high-precision 3D ball tracking in live football broadcasts using only a single, standard camera feed and no costly infrastructure.
\section{Related Works}
\label{ssec:related_taxonomy}

Existing approaches for three-dimensional ball localization can be categorized along two main axes: \emph{(i)~the number of cameras employed} and \emph{(ii)~the temporal mode of operation} (i.e. online vs. offline processing).
Table~\ref{tab:taxonomy} summarizes key publications within each category relevant to our work.

\begin{table}[h]
\centering
\caption{Literature taxonomy.}
\label{tab:taxonomy}
\setlength{\tabcolsep}{4pt}
\begin{tabular}{lccc}
\toprule
Method & Cameras & Time & Type \\ \midrule
Kim 1998~\cite{kim1998physics}           & 1   & offline  & parabola \\
Silva 2011~\cite{silva2012real}          & 1   & online   & EKF+ML \\
Shen 2015~\cite{shen20153d}                  & 1   & offline  & spline+ballistics \\
Van Zandycke 2022~\cite{van20223d}       & 1   & \underline{online}   & DL-diameter \\ 
Ribnick 2009~\cite{ribnick2008estimating} & 1   & offline   & fit trajectory \\
Metzler 2013~\cite{metzler20133d} & 1   & offline   & fit trajectory \\ 
\midrule
Ren 2008~\cite{ren2008real}          & 4   & \underline{near-time} & EKF \\
Maksai 2016~\cite{maksai2016players}              & 8   & offline   & MIP \\
\bottomrule
\end{tabular}
\end{table}

\subsubsection{Single-camera, offline approaches.}
Early solutions estimated depth by fitting a parabola to the observed 2D trajectory \cite{kim1998physics}. Ribnick \textit{et al.}\ \cite{ribnick2008estimating} developed this idea in detail, while Shen \textit{et al.}\ \cite{shen20153d} introduced spline interpolation with physical (ballistic) constraints for tracking balls in ping-pong. 
In their method, jump start and end positions are detected, and the trajectory is interpolated assuming rectilinear motion of the ball projection over the table -- an idea also applied to football, with different discrete ball states, by Strachan \textit{et al.}\ \cite{ren2008real}. 
Metzler \textit{et al.}\ \cite{metzler20133d} generated tracklets based on motion history images to reconstruct 3D trajectories using physical characteristics and camera calibration.

\subsubsection{Single-camera, online approaches.}
Silva \textit{et al.}\ \cite{silva2012real} dispensed with offline optimization windows, using an Extended Kalman Filter (EKF) and maximum likelihood estimation on short sequences. However, their method fails under extended ball occlusion.

More recently, approaches leveraging convolutional neural networks have emerged. For instance, Van Zandycke \textit{et al.}\ \cite{van20223d} employ a single-frame network to directly estimate ball diameter (and thus depth), achieving high speed but lacking velocity estimation or event detection as there is no physical modeling.

\subsubsection{Multi-camera, near-time and offline approaches.}
Ren \textit{et al.}\ \cite{ren2008real} presented a near-time algorithm using four static cameras, with a latency of \(\sim1-2\) seconds. 
Maksai \textit{et al.}\ \cite{maksai2016players} achieved high accuracy through global MIP optimization over 500 frames, at the cost of a much longer solution time (tens of seconds). 
Both works factorize the ball’s discrete state using latent variables.

\subsubsection{Positioning of the proposed method.}
Our algorithm combines the affordability and simplicity of single-camera setups with the low latency of online operation (down to a single frame), enhanced by a discrete event model. By doing so, we bridge the gap between fast yet limited single-camera EKF methods and the more accurate, but costly and complex, multi-camera algorithms leveraging latent state models.

\section{Problem Formulation}

We address the reconstruction of the three-dimensional trajectory of a football from a broadcast video. The inputs to this problem are detections of the ball and player positions on the field, together with the camera calibration parameters at each time step. Several factors complicate this task: the ball is frequently occluded by players, leading to frames with no visible ball; additionally, interactions with players and the environment, as well as abrupt trajectory changes, make accurate reconstruction challenging.

To overcome these difficulties, we explicitly model the interactions between the ball, players, and environment. Mathematically, the problem is cast as maximizing the posterior probability of the ball’s trajectory given the observed data. The solution is infered online by simulating possible ball states and pruning those with low probability.

\subsection{Input Data and Predictions}

For each frame at time step $t$, the algorithm receives as input: the positions of the ball on the screen $U_t = \{u_t^{(i)} \in \mathbb{R}^2\}$, the positions of the players on the field $G_t = \{g_{t}^{(i)} \in \mathbb{R}^2\}$, and the camera calibration $K_t, R_t, T_t$ parameters, where $K$ is the intrinsic camera matrix, $R$ is the rotation matrix, and $T$ is the translation vector. 
As output, for each frame $t$ our algorithm returns the most probable trajectory of the ball in space up to time $t$.

Denote the observed variables $\mathcal Z_{1:T}=\{U_{1:t},\,G_{1:t},\,K,\,R_{1:t},\,T_{1:t}\}$

\begin{equation}
  \label{eq:map}
  X_{1:T}^{\ast}
  \;=\;
  \arg\max_{X_{1:T}}
  \;p\!\bigl(X_{1:T}\mid\mathcal Z_{1:T}\bigr).
\end{equation}

\section{Online Solution}

Our method is implemented as a recursive filter featuring a hybrid continuous-discrete state and a limited search width within the trajectory space.
At each frame, we generate a finite set of hypotheses and assign each a log-likelihood score. Only the \(K\) most probable hypotheses are retained, with all others discarded. 
This approach ensures that the computational complexity scales linearly with the number of detections.

We reframe \eqref{eq:map} as a Markov model with hidden variables $H_t$:

$$
p(X_{1:t}, H_{1:t}\mid \mathcal{Z}_{1:t}) := p(X_t, H_t \mid X_{t-1}, H_{t-1}, \mathcal{Z}_t) \cdot p(X_{1:t-1}, H_{1:t-1}\mid \mathcal{Z}_{1:t-1})
$$

\subsection{Latent variables}

The  latent state of the filter $H_t$  includes a discrete mode \(s_t \in S = \{\mathcal{\mathcal{J}}, \mathcal{P}, \mathcal{W}, \mathcal{O}\}\), representing jumping, possession, waiting, and out-of-play states, along with additional parameters (see Sec.~\ref{sec:modes} for details).

\subsection{Log likelihood and pruning}

\label{ssec:beam}

We denote the pair $(X_t, H_t)$ as $\mathcal{X}_t$ and decompose the likelihood multiplier into three components:
$$
p_s(s_t+1 \mid s_t) \cdot p_x(X_{t+1} \mid X_t, s_{t+1}) \cdot p_z(\mathcal{Z}_{t+1}
                \mid X_{t+1}, s_{t+1})
$$
Here, $p_s$ is a hardcoded transition matrix precomputed from historical data. Computation of $p_x$ and $p_z$ is detailed in Sec.~\ref{sec:transitions}.

Following pruning at frame \(t\), we retain a set of hypotheses 
\(\{\mathcal{X}_t^{(n)}\}_{n=1}^{N_t}\)  
with corresponding accumulated log weights \(\ell_t^{(n)}\).
Each hypothesis generates a set of descendants  
\(\{\mathcal{X}_{t+1}^{(n,m)}\}\):
the index \(n\) indicates the \emph{parent},  
while \(m\) indexes a specific \emph{descendant} in frame \(t\!+\!1\).
The log-likelihood increment for a transition  
\(\mathcal{X}_t^{(n)} \rightarrow \mathcal{X}_{t+1}^{(n,m)}\) is given by
\[
  \Delta\ell_{t+1}^{(n,m)} =
    \underbrace{
      \log p_s\bigl(s_{t+1}^{(n, m)}\mid s_t^{(n)}\bigr)
      + \log p_x\bigl(X_{t+1}^{(n, m)}\mid X_t^{(n)}, s_{t+1}^{(n, m)}\bigr)
      }_{\text{dynamics}} +
\]
\[
   + \underbrace{\log
      p_z\bigl(\mathcal{Z}_{t+1}
                \mid X_{t+1}^{(n, m)}, s_{t+1}^{(n, m)}\bigr)}_{\text{observation}}
\]

\noindent
The total weight for each descendant is recursively computed as 
\(\ell_{t+1}^{(n,m)} = \ell_t^{(n)} + \Delta\ell_{t+1}^{(n,m)}\).
After evaluating all descendants,
merge the resulting set,
sort by \(\ell_{t+1}^{(n,m)}\) in descending order,
and retain the top \(K\) hypotheses,
discarding the remainder.

\subsection{Chain Extraction with an Online Buffer of Size $L$}
\label{ssec:buffer}
At each frame, the above algorithm produces a set of candidate hypotheses $\{\mathcal{X}_{t}^{(n)}\}$. We select the hypothesis with the highest log-likelihood $\ell_{t}^{(n)}$ and then recursively trace back through its chain of ancestors to recover the trajectory $X_{1:t}$.

However, only the part of the trajectory up to frame \(t-L\) is output immediately;  
the most recent \(L\) estimates remain \emph{soft}  
and can be revised as new observations are received.  
This buffering enables dynamic control over the algorithm’s latency and improves prediction accuracy when immediate frame-by-frame estimates are not required.

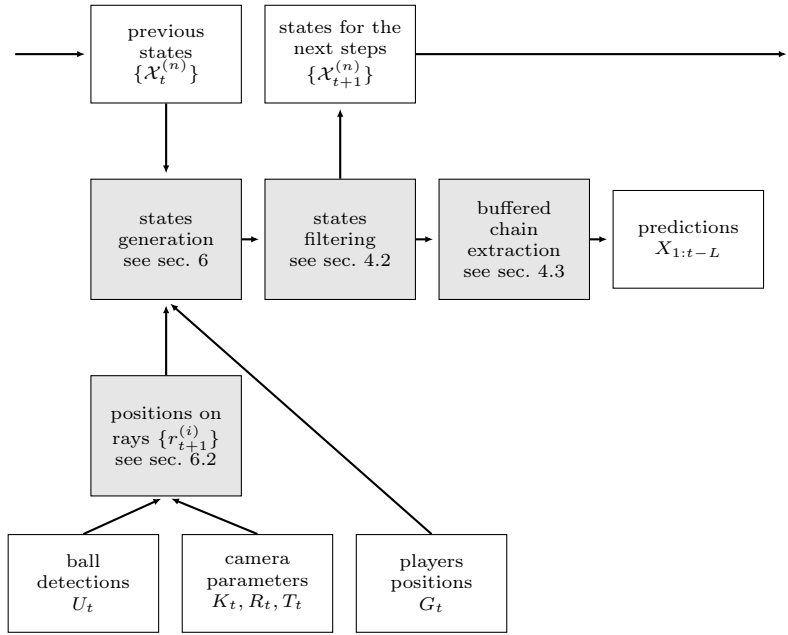
\begin{figure}[ht]
    \centering
    \begin{tikzpicture}[
  node distance=1cm and 0.3cm,
  process/.style = {draw, rectangle, fill=gray!20,
                    minimum width=2.0cm,  
                    minimum height=1.6cm, 
                    align=center, font=\scriptsize},
  data/.style    = {draw, rectangle, fill=white,
                    minimum width=2.0cm,
                    minimum height=1.3cm,
                    align=center, font=\scriptsize},
  >={Latex[length=3pt,width=2.5pt]},
  arr/.style     = {->, line width=.8pt, shorten >=2pt}
]

  \node[data] (prev) {previous\\states\\$\{\mathcal{X}_t^{(n)}\}$};

  \node[process] (gen) [below=1cm of prev] {states\\generation\\see sec.~\ref{sec:transitions}};
  \node[process] (fil) [right=of gen]         {states\\filtering\\see sec.~\ref{ssec:beam}};
  \node[data] (next)   [above=of fil] {states for the\\next steps\\$\{\mathcal{X}_{t+1}^{(n)}\}$};
  \node[process] (buf) [right=of fil]         {buffered\\chain\\extraction\\see sec.~\ref{ssec:buffer}};
  \node[data]    (pred)[right=of buf]         {predictions\\$X_{1:t-L}$};

  \node[process] (pos) [below=of gen] {positions on\\rays $\{r_{t+1}^{(i)}\}$\\see sec.~\ref{sec:ray_hypotises}};

  \node[data] (ball) [below=of pos, xshift=-1.1cm, yshift=0.5cm] {ball\\detections\\$U_t$};
  \node[data] (cam)  [right=of ball]                 {camera\\parameters\\$K_t, R_t, T_t$};
  \node[data] (play) [right=of cam]  {players\\positions\\$G_t$};

  \draw[arr] ([xshift=-1cm]prev.west) -- (prev.west);
  \draw[arr] (prev.south) -- (gen.north);

  \draw[arr] (gen.east) -- (fil.west);
  \draw[arr] (fil.east) -- (buf.west);
  \draw[arr] (buf.east) -- (pred.west);

  \draw[arr] (fil.north) -- (next.south);
  \draw[arr] (next.east) -- ++(5cm,0); 

  \draw[arr] (ball.north) -- (pos.south);
  \draw[arr] (cam.north)  -- (pos.south);
  \draw[arr] (pos.north)  -- (gen.south);

  \draw[arr] (play.north) -- (gen.south);

\end{tikzpicture}
    \caption{The scheme of the proposed pipeline.}
    \label{scheme}
\end{figure}

\section{Ball State Modes}

\subsection{Discrete Modes of Ball Motion}
\label{sec:modes}

To model the variety of game events, we define a finite set of discrete ball states
\(
S=\{\mathbf{\mathcal J},\mathcal P,\mathcal W,\mathcal O\}.
\)
For each frame~\(t\) the discrete variable \(s_t\), included in the latent state \(H_t\), specifies which kinematic model applies.

\subsubsection{\(\mathbf{\mathcal J}\) — Jumping}
In the jumping mode, the ball is not in contact with any player and moves purely under the force of gravity, potentially bouncing on the field.
No player interaction occurs in this state.
In addition to position, \(\mathcal{J}\)-states also store \(v_t\) (the current ball velocity) and \(\Sigma_t\) (the position variance). J-states may be visible or occluded (blocked or undetected). The algorithm for generating J-states is detailed in Section~\ref{sec:transitions}.

\subsubsection{\(\mathcal P\) — Player Possession}
In possession mode, the ball’s center is aligned with a player’s coordinates and its independent motion is suppressed. Player positions generate corresponding  P-states, producing one for each player per frame.

\subsubsection{\(\mathcal W\) — Wait-after-Possession}
The waiting mode is a buffer state after the ball leaves a player's control but before it is visually detected, supporting robust initialization of the jumping phase. 
Besides position, \(\mathcal{W}\)-state maintains \(b_t\) (the frame when the ball was kicked) and \(g_t\) (the coordinates of the kicker).

While \(s_t=\mathcal{W}\), the filter stores only the kick time and 2-D player coordinates, \textbf{generating no ballistic hypotheses}. 
Only when the ball first becomes visible
\(u_{t}^i\) appears, possible positions inferred from the camera’s viewing ray and the state transitions to  \(\mathbf{\mathcal J}\) mode.
This design ensures that ballistic calculations are run only once per appearance, rather than on every possession frame -- dramatically reducing the number of required hypotheses, maintaining real-time beam width up to \(K\approx 1000\). 

The W-states are generated from the previous frame’s W-states or from current P-states.

\subsubsection{\(\mathcal O\) — Out-of-Pitch}
The out-of-pitch state models periods when the ball is outside the field, awaiting to reenter the game (e.g., throw-in or corner kick). No extra latent variables are stored in this mode. The O-states are generated from the current frame J-states found to be outside the soccer pitch boundaries.

\section{State Transitions}
\label{sec:transitions}

\begin{figure}[ht]
    \centering
    \begin{tikzpicture}[
  node distance=1cm,
  state/.style   = {circle,draw,thick,minimum size=18pt,inner sep=0pt},
  >={Latex[length=3pt,width=6pt]}
]
  \node[state] (J) {J};
  \node[state] (O) [right=of J] {O};
  \node[state] (W) [below=of J] {W};
  \node[state] (P) [below=of O] {P};

  \draw[thick,->,loop left]  (J) to (J);
  \draw[thick,->,loop right] (O) to (O);
  \draw[thick,->,loop left]  (W) to (W);
  \draw[thick,->,loop right] (P) to (P);

  \draw[thick,->] (J) -- (O);   
  \draw[thick,->] (O) -- (P);   
  \draw[thick,->] (P) -- (W);   
  \draw[thick,->] (W) -- (J);   
  \draw[thick,->] (J) -- (P);   
\end{tikzpicture}
    \caption{Possible types of state transitions.}
    \label{transitions}
\end{figure}
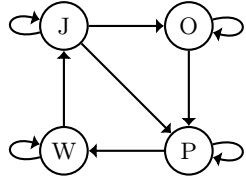

All possible transitions between ball states over time are illustrated in Fig~\ref{transitions}. 
Some transitions, such as $\mathcal{O}\to \mathcal{O}$, $\mathcal{W}\to \mathcal{W}$, $\mathcal{J}\to \mathcal{O}$, $\mathcal{P}\to \mathcal{W}$ are handled by simply inheriting the position from the previous state.
In these cases, the new state's position $X_t$ is copied directly from the prior state. The log likelihood increment $\Delta \ell$ depends solely on the transition probability $p_s$ and reduces to $\Delta \ell=\log p_s(s_{t+1}\mid s_t)$.

\subsection{$\mathrm{any} \to \mathcal {P}$ transitions}

For transitions into player possession ($\mathcal{P}$-states), the log-likelihood increment $\Delta \ell$  additionally uses the function $p_x$, which can only take values 0 or 1, determined by the distance between $X_{t+1}$ and $X_t$:
$$
p_x(X_{t+1} \mid X_t, s_{t+1}=\mathcal{P}) = \|X_{t+1} - X_{t}\| < \mathrm{thrsld}
$$
Here, $\mathrm{thrsld}$ is a distance threshold precomputed from historical data. Thus, $\Delta \ell=\log p_s(s_{t+1}=\mathcal{P}\mid s_t)$ if the new position $X_{t+1}$ is sufficiently close to the previous one; otherwise, $\Delta \ell = -\infty$, effectively eliminating improbable transitions.

\subsection{Hypothesis Generation from Visible Detections}
\label{sec:ray_hypotises}

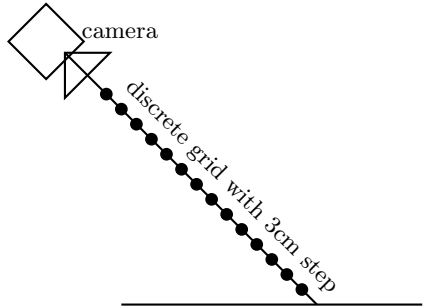
\begin{figure}[ht]
    \centering
    \begin{tikzpicture}[scale=2]

  \draw[thick] (0.5,0.0) -- (2.5,0.0);
  \draw[thick] (0.0,2.0) -- (0.25, 1.75) -- (0.0, 1.5) -- (-0.25, 1.75) -- cycle;
  \draw[thick] (0.125, 1.675) -- (0.425, 1.675) -- (0.125, 1.375) -- cycle;
  \node[black, below right] at (0.18, 1.9) {\footnotesize camera};
  \draw[thick] (0.125, 1.675) -- (1.8,0.0);
  \draw[fill=black, draw=black, thick] (0.4, 1.4) circle (1pt);
  \draw[fill=black, draw=black, thick] (0.5, 1.3) circle (1pt);
  \draw[fill=black, draw=black, thick] (0.6, 1.2) circle (1pt);
  \draw[fill=black, draw=black, thick] (0.7, 1.1) circle (1pt);
  \draw[fill=black, draw=black, thick] (0.8, 1) circle (1pt);
  \draw[fill=black, draw=black, thick] (0.9, 0.9) circle (1pt);
  \draw[fill=black, draw=black, thick] (1,   0.8) circle (1pt);
  \draw[fill=black, draw=black, thick] (1.1, 0.7) circle (1pt);
  \draw[fill=black, draw=black, thick] (1.2, 0.6) circle (1pt);
  \draw[fill=black, draw=black, thick] (1.3, 0.5) circle (1pt);
  \draw[fill=black, draw=black, thick] (1.4, 0.4) circle (1pt);
  \draw[fill=black, draw=black, thick] (1.5, 0.3) circle (1pt);
  \draw[fill=black, draw=black, thick] (1.6, 0.2) circle (1pt);
  \draw[fill=black, draw=black, thick] (1.7, 0.1) circle (1pt);

    \node[black, below right, rotate=-45] at (0.6, 1.6) {\footnotesize discrete grid with 3cm step};

\end{tikzpicture}
    \caption{Generated positions for visible J-states}
    \label{ray2}
\end{figure}

Given the camera calibration parameters and ball detection in the image, the ball’s 3D location cannot be uniquely determined; instead, possible positions are constrained to a ray originating from the camera point. For every detection, we discretize this ray in 3-centimeter increments, from the field surface up to the camera’s position, and generate candidate J-states (see Fig.~\ref{ray2}). We denote these visible positions as $\{r_t^{(i)}\}$.

\subsection{$\mathcal {J} \to \mathcal {J}$ transition}

When in jumping mode, the current $\mathcal{J}$-state generates both visible and invisible $J$-states, depending on whether a detection is associated with it. First, the invisible state is created by extrapolating the previous state’s dynamics over a time step  $\mathrm{dt}$, where $\mathrm{dt}$ is the time between consecutive frames: $\mathrm{dt}=1/\mathrm{fps}$.
Using the previous spatial position and the velocity vector, the new position is computed.
We introduce the extrapolation function $j(X_t, v_t, \mathrm{dt})$, that yields the updated position and velocity vector $X'_{t+1}, v_{t+1}$.  
To compute the likelihood of visible states we additionally require the spatial position variance, which describes the uncertainty in the current position. We update the variance as follows: $\Sigma'_{t+1} = \Sigma_t + \mathrm{dt} \cdot \Sigma_v$, where $\Sigma_v$ is derived from historical data.
The increment of the log-likelihood is then given by $\Delta \ell = \log p_s(s_{t+1}=\mathcal{J} \mid s_t=\mathcal{J}) + \log p_z^{\mathrm{invisible}}$, where $p_z^{\mathrm{invisible}}$ is a pre-computed value used to penalize the state for the absence of detection.

To generate visible hypotheses, we associate invisible candidates with available detections. 
For each discrete ray position $r_{t+1}^{(i)}$ (see Sec.~\ref{sec:ray_hypotises}) a new visible hypothesis is formed: $X_{t+1}^{\mathrm{visible}}:=r_{t+1}^{(i)}$. The likelihood component $p_x$ is computed as the value of a of a Gaussian with mean $X'_{t+1}$ and variance $\Sigma'_{t+1}$ at $X_{t+1}=r_{t+1}^{(i)}$, i.e. $p_x = N(r_{t+1}^{(i)} \mid X'_{t+1}, \Sigma'_{t+1})$. 
This formulation penalizes trajectories that deviate from physical plausibility.
Hence, the better the ball’s trajectory follows the laws of physics, the smaller the penalty. The factor $p_z$ is set to 1, $\Sigma_{t+1}$ is set to 0.

\subsection{$\mathcal {W} \to \mathcal {J}$ transition}

For each visible position on the ray $r_{t+1}^{(i)}$ (see Section~\ref{sec:ray_hypotises}) and $\mathcal{W}$ -- a new $J$-state is initialized, with position $X_{t+1} = r_{t+1}^{(i)}$. The velocity vector is computed with the free-flight equation, using the ball’s initial position $X_t$, final position $X_{t+1}$, and the time the ball spent without detection (this time is part of the hidden state of type $\mathcal{W}$). At the first glance, the variance $\Sigma_{t+1}$ of the new $J$-state should be minimal, because we have explicitly modelled its current position; however, we assign it a predetermined value $\Sigma_g$.  
While the new position is precisely modeled by the detection, its point of origin (the exact kick location) is uncertain—since only the player's position at the kick moment is observed, and the part of the body used to strike the ball can vary (flight direction significantly depends on which body part was used).
Rather than introducing the variance of the velocity vector, we assign the described uncertainty to $\Sigma_{t+1}$, thereby reducing the number of latent variables of the $J$-state. 
This approach is mathematically consistent, as subsequent computations rely only on combined variances of position and velocity.

Here, the increment of the log-likelihood solely depends on $p_s(s_{t+1}=\mathcal{J} \mid s_t=\mathcal{W})$.

\section{Experiments}

\subsection{Data}
\label{ssec:dataset}

For quantitative analysis, we used a dataset collected from broadcast footage of the Russian Premier League 2024/25 season. 
The sample comprises 3-minute segments from ten matches (a total of 30 minutes). 
The source video was captured at a resolution of $6144\times3240$ pixels and $25$ frames per second. 
For each frame, the ball’s ground-truth position was obtained via multi-view triangulation using four calibrated cameras placed around the perimeter of the stadium.
The mean triangulation error, measured using control markers, is $<\!10$ cm.
The discrete modes
\(\mathbf{\mathcal J},\mathcal P,\mathcal W,\mathcal O\)
were labeled manually. The pixel coordinates of the ball in the broadcast view were obtained using the RTMDet-t detector. Player detections were obtained with the same detector, then projected onto the pitch plane using homography.

\subsection{Baseline Methods}
\label{ssec:baselines}

To evaluate the proposed online filter, we selected a set of monocular algorithms representing different generations of approaches.

\begin{enumerate}
  \item \textbf{Kim 1998}~\cite{kim1998physics}\;
        — approximation of the 2-D trajectory with a parabola and recovery of depth based on the throw parameters.
        Operates offline over the entire episode with a single run of a nonlinear least squares solver.

  \item \textbf{Metzler 2013}~\cite{metzler20133d}\;
        — 3-D reconstruction based on the \emph{Motion History Image} technique:
        segmentation of the ball’s orbits in 2-D, followed by fitting a physical trajectory;
        designed for static cameras, operates offline.

  \item \textbf{Ribnick 2009}~\cite{ribnick2008estimating}\;
        — analytical EKF that derives conditions for the unique recovery of the projectile’s depth
        and velocity from a single camera; does not take into account the game scene or player interactions.

  \item \textbf{Maksai 2016}~\cite{maksai2016players}\;
        — multi-camera MIP optimization over a 500-frame window;
        included as an offline upper bound for accuracy, but not comparable in speed and latency.
\end{enumerate}

\subsection{Evaluation Metrics}
\label{ssec:metrics}

\subsubsection{Localization Accuracy.}
We use a threshold metric
\emph{Accuracy@\(d\)} — the fraction of frames
in which the positional error
does not exceed a given threshold \(d\).

Results are reported for
\(d\in\{0.5,\;1,\;2,\;4,\;8\}\,\mathrm{m}\).
This scale provides both
"tele-accuracy" ($\leq$ 50 cm, suitable for
expected goals analytics)
and coarse localization ($\leq$ 8 m),
sufficient for visualization in TV broadcasts.

\subsubsection{Game Event Accuracy.}
For key discrete transitions: \textit{kick}
and \textit{out-of-pitch} (OOP) \(F_1\) score are computed
using a tolerance window
of \(\pm\,12\) frames around the GT label.

\subsubsection{Speed and Latency.}
\emph{Throughput} (fps) is the number of frames processed per second.
\emph{Latency} is defined as
the difference between the moment a frame becomes available and
the moment the algorithm outputs
an estimate for that frame, averaged over the test set
(expressed in frames).
For offline methods, the total time for processing the clip is given,
converted to equivalent fps,
and the latency is set equal to the clip length.
To analyze the accuracy-latency trade-off, our algorithm is run with an adjustable latency buffer parameter \(0 \leq L \leq 50\) frames,
as described in Sec.~\ref{ssec:beam}.

\begin{table*}[t]
\centering
\caption{Comparison of accuracy and speed.
Best results are in bold, second best are underlined.}
\label{tab:results}
\setlength{\tabcolsep}{4pt}
\begin{tabular}{lccccccccc}
\toprule
\multirow{2}{*}{Method} &
\multicolumn{5}{c}{Accuracy@\,$d$ (m)} & \multicolumn{2}{c}{$F_1$ events} &
\multirow{2}{*}{fps} & \multirow{2}{*}{Latency}\\
\cmidrule(lr){2-6}\cmidrule(lr){7-8}
& 0.5 & 1 & 2 & 4 & 8 & kick & OOP & & \\
\midrule
Ours, $L{=}1$  & 0.50 & 0.61 & 0.65 & 0.69 & 0.71 & 0.59 & 0.64 & {53} & 1 \\
Ours, $L{=}10$          & 0.56 & 0.63 & 0.66 & 0.69 & 0.71 & 0.61 & 0.66 & {53} & 10 \\
Ours, $L{=}25$          & \underline{0.59} & 0.64 & 0.67 & 0.70 & 0.71 & 0.71 & 0.72 & {53} & 25 \\
Ours, $L{=}50$          & \textbf{0.59} & \textbf{0.66} & \underline{0.67} & \underline{0.71} & \underline{0.71} & \textbf{0.74} & \underline{0.79} & {53} & 50 \\
\midrule
Kim 1998        & 0.21 & 0.39 & 0.52 & 0.61 & 0.68 & --- & --- & 9 & 4\,500$^{\dagger}$ \\
Metzler 2013         & 0.34 & 0.49 & 0.59 & 0.64 & 0.69 & --- & --- & 6  & 4\,500$^{\dagger}$ \\
Ribnick 2009         & 0.30 & 0.56 & 0.63 & 0.67 & 0.69 & --- & --- & 21 & 4\,500$^{\dagger}$ \\
\midrule
Maksai 2016 (MIP)      & {0.56} & \underline{0.64} & \textbf{0.68} & \textbf{0.73} & \textbf{0.77} & \underline{0.66} & \textbf{0.81} & 6 & 500$^{\dagger}$ \\
\bottomrule
\end{tabular}

\vspace{2pt}
\footnotesize
$^{\dagger}$For offline methods, latency is equal to the length of the three-minute clip  
($3\,\mathrm{min}\times25\,\mathrm{fps}=4\,500$ frames);
fps is computed as the number of frames divided by the total processing time.
\end{table*}

\subsection{Results Analysis}
\label{ssec:discussion}

Table~\ref{tab:results} shows that
with a minimal latency of \(L{=}1\) frame,
our filter is inferior to the offline
Maksai~2016 (multi-camera MIP) method at all accuracy thresholds.
However, increasing the latency to \(L{=}25\) nearly closes the accuracy gap,
and at \(L{=}50\) frames
our online solution becomes
\emph{comparable} to or even
\emph{surpasses} the MIP baseline in several respects:

\begin{itemize}
  \item At the tele-accuracy levels
        \(\text{Acc}_{0.5}\) and \(\text{Acc}_{1}\)
        our filter outperforms MIP
        by \(+0.03\) and \(+0.02\), respectively
        (0.59 vs 0.56 and 0.66 vs 0.64);
  \item For game events,
        a higher $F_1$ for \textit{kick}
        is achieved (0.74 versus 0.66),
        which is particularly valuable for live analytics of shots;
  \item  Our approach maintains real-time throughput( \(50\)~fps) even at the highest latency, in contrast to the substantially slower MIP method
        (\(6\)~fps).
\end{itemize}

The advantage of our algorithm stems from its tailored design for football:
the discrete modes \(\mathcal P,\mathcal W,\mathcal O\)
explicitly model possession, kicking, and out-of-bounds events, and the \(\mathcal W\) buffering defers costly ballistic updates until reliable detections appear.
Maksai’s general-purpose MIP solver does not distinguish these micro-events and thus accumulates error in crowded or occluded scenes.

In summary, for practical latencies
\(L\leq50\) frames (2 seconds at 25~fps),
our online filter achieves accuracy
that matches or exceeds offline optimization,
while delivering an order-of-magnitude reduction in computational cost (\(\sim\!10\) time reduction of the inference time).

\section{Conclusion}

In this work, we have presented a novel multi-mode state model specifically tailored for the requirements of football broadcast analytics. By structuring the filter to operate on a finite set of discrete and interpretable ball states (Jumping ($\mathcal{J}$), Player Possession ($\mathcal{P}$), Wait-after-Possession ($\mathcal{W}$), and Out-of-Pitch ($\mathcal{O}$)), we achieve both efficient computation and domain-specific modeling. This design allows us to capture micro-events such as kicks and possession changes, while the introduction of the W-buffer efficiently bridges the moment between a kick and the first detection of the ball. As a result, the computational complexity associated with ballistic trajectory hypotheses is drastically reduced.

Our experimental evaluation, conducted on high-resolution broadcast footage from the Russian Premier League, demonstrates that at moderate latency settings ($L \leqslant 50$ frames), the proposed online filter reaches or surpasses the localization and event-detection accuracy of the state-of-the-art offline optimization method by Maksai et al., but at approximately one-tenth the computational cost. Notably, our filter achieves superior tele-accuracy (Acc0.5, Acc1) and higher F1 for critical game events such as kicks, while maintaining true real-time throughput (50 fps).

These results underscore the benefits of integrating domain knowledge into state estimation pipelines for sports analytics. The proposed approach enables fast, accurate, and interpretable analysis of broadcast video, facilitating a range of applications from live sports analytics to TV visualization. In future work, we plan to enrich the set of modeled football events and investigate robust deployment strategies for diverse broadcast and analytic infrastructures.

%
%
%
\bibliographystyle{splncs04}
\bibliography{mybibliography}
\end{document}